\documentclass[a4,a4wide]{article}

\usepackage{aaai}
\usepackage{times}
\usepackage{helvet}
\usepackage{courier}
\usepackage{amsmath,amssymb,times,xspace,algorithmic,graphicx}

\usepackage[caption=false]{subfig}

\usepackage{listings}
\usepackage[usenames,dvipsnames]{xcolor}

\newcounter{definitionnumber}

\usepackage{mdwlist}

\newcommand{\dmh}{\rho}

\newcommand{\PrASP}{PrASP }

\newcommand{\kb}{\Lambda}

\newcommand{\as}{\Gamma}

\newcommand{\folTolp}{\mathit{lp}}

\begin{document}  
\nocopyright


\setcounter{secnumdepth}{1}  

\title{Probabilistic Inductive Logic Programming Based on Answer Set Programming\thanks{This work is an extended and revised version of A. Mileo, M. Nickles: Probabilistic Inductive Answer Set Programming by Model Sampling and Counting. First International Workshop on Learning and Nonmonotonic Reasoning (LNMR 2013), Corunna, Spain, 2013.\newline}}

\author{ Matthias Nickles$^{\sharp\flat}$ and Alessandra Mileo$^{\sharp}$\\
 \texttt{\{matthias.nickles,alessandra.mileo\}@deri.org}\\
 $^{\sharp}$ INSIGHT/DERI Galway\\
 National University of Ireland, Galway\\
$^{\flat}$ Department of Information Technology\\
 National University of Ireland, Galway\\ 
}

\maketitle

\begin{abstract}

We propose a new formal language for the expressive representation of probabilistic knowledge based on Answer Set Programming (ASP). It allows for the annotation of first-order formulas as well as ASP rules and facts with probabilities and for learning of such weights from data (parameter estimation). Weighted formulas are given a semantics in terms of soft and hard constraints which determine a probability distribution over answer sets. In contrast to related approaches, we approach inference by optionally utilizing so-called streamlining XOR constraints, in order to reduce the number of  computed answer sets. Our approach is prototypically implemented. Examples illustrate the introduced concepts and point at issues and topics for future research.\\ 


\end{abstract}

\section{Introduction}
\label{sec:intro}
 
Reasoning in the presence of uncertainty and relational structures (such as social networks and Linked Data) is an important aspect of knowledge discovery and representation for the Web, the Internet Of Things, and other potentially heterogeneous and complex domains. Probabilistic logic programing, and the ability to learn probabilistic logic programs from data, can provide an attractive approach to uncertainty reasoning and statistical relational learning, since it combines the deduction power and declarative nature of logic programming with probabilistic inference abilities traditionally known from less expressive graphical models such as Bayesian and Markov networks. A very successful type of logic programming for nonmonotonic domains is \textit{Answer Set Programming} (ASP) \cite{answer-set-programming,stable-model-semantics}. Since statistical-relational approaches to probabilistic reasoning often rely heavily on the propositionalization of first-order or other relational information, ASP appears to be an ideal basis for probabilistic logic programming, given its expressiveness and the existence of highly optimized grounders and solvers. However, despite the successful employment of conceptually related approaches in the area of SAT  for probabilistic inference tasks, only a small number of  approaches to probabilistic knowledge representation or probabilistic inductive logic programming under the stable model semantics exist so far, of which some are rather restrictive wrt. expressiveness and parameter estimation techniques. We build upon these and other existing approaches in the area of probabilistic (inductive) logic programming in order to provide a new ASP-based probabilistic logic programming language (with first-order as well as ASP basic syntax) for the representation of probabilistic knowledge. Weights which directly represent probabilities can be attached to arbitrary formulas, and we show how this can be used to perform probabilistic inference and how weights of hypotheses can be inductively learned from given relational examples. To the best of our knowledge, this is the first ASP-based approach to probabilistic (inductive) logic programming which does not impose restrictions on the annotation of ASP-rules and facts as well as FOL-style formulas with probabilities. 

The remainder of this paper is organized as follows: the next section presents relevant related approaches. Section 3 introduces syntax and semantics of our new language. Section 4 presents our approach to probabilistic inference (including examples), and Section 5 shows how formula weights can be learned from data. Section 6 concludes.

\section{Related Work}
\label{sec:related}
 
Being one of the early approaches to the logic-based representation of uncertainty sparked by Nilsson's seminal work \cite{nilsson}, \cite{halpern90} presents three different probabilistic first-order languages, and compares them with a related approach by Bacchus \cite{bacchus90}. One language has a \emph{domain-frequency} (or \textit{statistical}) semantics, one has a possible worlds semantics (like our approach), and one bridges both types of semantics. While those languages as such are mainly of theoretical relevance, their types of semantics still form the backbone of most practically relevant contemporary approaches.\\   
Many newer approaches, including Markov Logic Networks (see below), require a possibly expensive grounding (propositionalization) of first-order theories over finite domains. A recent approach which does not fall into this category but employs the principle of maximum entropy in favor of performing extensive groundings is \cite{journals/igpl/ThimmK12}. However, since ASP is predestined for efficient grounding, we do not see grounding necessarily as a shortcoming. \emph{Stochastic Logic Programs} (SLPs) \cite{muggleton00} are an influential approach where sets of rules in form of range-restricted clauses can be labeled with probabilities. Parameter learning for SLPs is approached in \cite{Cussens00parameterestimation} using the EM-algorithm. 
Approaches which combine concepts from Bayesian network theory with relational modeling and learning are, e.g., \cite{Friedman99learningprobabilistic,kersting00,laskey05}. Probabilistic Relational Models (PRM) \cite{Friedman99learningprobabilistic} can be seen as relational counterparts to Bayesian networks 
In contrast to these, our approach does not directly relate to graphical models such as Bayesian or Markov Networks but works  on arbitrary possible worlds which are generated by ASP solvers. ProbLog \cite{DBLP:conf/ijcai/RaedtKT07} allows for probabilistic facts and definite clauses, and approaches to probabilistic rule and parameter learning (from interpretations) also exist for ProbLog. Inference is based on weighted model counting, which is similarly to our approach, but uses Boolean satisfiability instead of stable model search. ProbLog builds upon the very influential Distribution Semantics introduced for PRISM \cite{Sato97prism:a}, which is also used by other approaches, such as  Independent Choice Logic (ICL) \cite{Poole97theindependent}. Another important approach outside the area of ASP are \emph{Markov Logic Networks} (MLN) \cite{mln}, which are related to ours. A MLN consists of first-order formulas annotated with weights (which are not probabilities). MLNs are used as ``templates'' from which Markov networks are constructed, i.e., graphical models for the joint distribution of a set of random variables. The (ground) Markov network generated from the MLN then determines a probability distribution over possible worlds. MLNs are syntactically similar to the logic programs in our framework (in our framework, weighted formulas can also be seen as soft or hard constraints for possible worlds), however, in contrast to MLN, we allow for probabilities as formula weights. Our initial approach to weight learning is closely related to certain approaches to MLN parameter learning (e.g., \cite{Lowd07efficientweight}), as described in Section 5.\\
Located in the field of nonmonotonic logic programming, our approach is also influenced by P-log \cite{DBLP:journals/corr/abs-0812-0659} and abduction-based rule learning in probabilistic nonmonotonic domains \cite{Corapi:2011:PRL:2044543.2044565}. With P-log, our approaches shares the view that answer sets can be seen as possible worlds in the sense of \cite{nilsson}. However, the syntax of P-log is quite different from our language, by restricting probabilistic annotations to certain syntactical forms and by the concept of independent experiments, which simplifies the implementation of their framework. In distinction from P-log, there is no particular coverage for causality modeling in our framework. \cite{Corapi:2011:PRL:2044543.2044565} allows to associate probabilities with abducibles and to learn both rules and probabilistic weights from given data (in form of literals). In contrast, our present approach does not comprise rule learning. However, our weight learning algorithm allows for learning from any kind of formulas and for the specification of virtually any sort of hypothesis as learning target, not only sets of abducibles. Both \cite{Corapi:2011:PRL:2044543.2044565} and our approach employ gradient descent for weight learning. Other approaches to probabilistic logic programming based on the stable model semantics for the logic aspects include \cite{Saad05hybridprobabilistic} and \cite{DBLP:journals/iandc/NgS94}. \cite{Saad05hybridprobabilistic} appears to be a powerful approach, but restricts probabilistic weighting to certain types of formulas, in order to achieve a low computational reasoning complexity. Its probabilistic annotation scheme is similar to that proposed in \cite{DBLP:journals/iandc/NgS94}. \cite{DBLP:journals/iandc/NgS94} provides both a language and an in-depth investigation of the stable model semantics (in particular the semantics of non-monotonic negation) of probabilistic deductive databases.\\ 
Our approach (and ASP in general) is closely related to SAT solving, \#SAT and constraint solving. ASP formulas in our language are constraints for possible worlds (legitimate models). As \cite{DBLP:conf/aaai/SangBK05} shows, Bayesian networks can be ``translated'' into a weighted model counting problem over propositional formulas, which is related to our approach to probabilistic inference, although details are quite different. Also, the XOR constraining approach \cite{DBLP:conf/nips/GomesSS06} employed for sampling of answer sets (Section \ref{Sampling}) has originally been invented for the sampling of propositional truth assignments. 

\section{Probabilistic Answer Set Programming with \PrASP}
\label{\PrASP}

Before we turn to probabilistic inference and parameter estimation, we introduce our new language for probabilistic non-monotonic logic programming, called Probabilistic Answer Set Programming (\textit{\PrASP}). 

\subsection{Syntax: Just add probabilities}
\label{syntax}

To remove unnecessary syntax restrictions and because we will later require certain syntactic modifications of given programs which are easier to express in First-Order Logic (FOL) notation, we allow for FOL statements in our logic programs, using the F2LP conversion tool \cite{conf/lpnmr/LeeP09}. More precisely, a \textit{\PrASP program} 
consists of ground or non-ground formulas in unrestricted first-order syntax annotated with numerical \textit{weights} (provided by some domain expert or learned from data). Weights directly represent  probabilities. If the weights are removed, and provided finite variable domains, any such program can be converted into an equivalent answer set program by means of the transformation described in \cite{conf/lpnmr/LeeP09}. 

Let $\varPhi$ be a set of function, predicate and object symbols and $\mathcal{L(\varPhi)}$ a first-order language over $\varPhi$ and the usual connectives (including both strong negation ``-'' and default negation ``not'') and first-order quantifiers.\\  
 Formally, a \PrASP program is a non-empty finite set $\{ ([p], f_i)\}$ of PrASP \textit{formulas} where each formula $f_i \in \mathcal{L(\varPhi)}$ is annotated with a \textit{weight} $[p]$. A weight directly represents a probability (provided it is probabilistically sound). If the weight is omitted for some formula of the program, weight $[1]$ is assumed. The weight $p$ of [$p$] $f$ is denoted as $w(f)$. Weighted formulas can intuitively seen as constraints which specify which possible worlds are indeed possible, and with which probability.\\
 Let $\kb^-$ denote PrASP program $\kb$ stripped of all weights. Weights need to be probabilistically sound, in the sense that the system of inequalities (\ref{linearSystem}) - (\ref{linearSystem4}) in Section \ref{semantics} must have at least one solution (however, in practice this does not need to be strictly the case, since the constraint solver employed for finding a probability distribution over possible worlds can find approximate solutions often even if the given weights are inconsistent).

In order to translate conjunctions of unweighted formulas in first-order syntax into disjunctive programs with a stable model semantics, we further define transformation $\folTolp: \mathcal{L(\varPhi)} \cup dLp(\varPhi) \rightarrow dLp(\varPhi)$, where $dLp(\varPhi)$ is the set of all disjunctive programs over $\varPhi$. The details of this transformation can be found in \cite{conf/lpnmr/LeeP09}\footnote{The use of the translation into ASP syntax requires either an ASP solver which can deal directly with disjunctive logic programs (such as claspD) or a grounder which is able to shift disjunctions from the head of the respective rules into the bodies, such as gringo \cite{clasp}.}. Applied to rules and facts in ASP syntax, $\folTolp$ simply returns these. This allows to make use of the wide range of advanced possibilities offered by contemporary ASP grounders in addition to FOL syntax (such as aggregates), although when defining the semantics of programs, we consider only formulas in FOL syntax. 

\subsection{Semantics}
\label{semantics}

The probabilities attached to formulas in a \PrASP program induce a probability distribution over answer sets of an ordinary answer set program which we call the \textit{spanning program} associated with that \PrASP program. Informally, the idea is to transform a \PrASP program into an answer set program whose answer sets reflect the nondeterminism introduced by the probabilistic weights: each annotated formula might hold as well as not hold (unless its weight is [0] or [1]). Of course, this transformation is lossy, so we need to memorize the weights for the later computation of a probability distribution over possible worlds. The important aspect of the spanning program is that it programmatically generates a set of possible worlds in form of answer sets. \\
Technically, the spanning program $\dmh(\kb)$ of PrASP program $\kb$ is a disjunctive program obtained by transformation $\folTolp(\kb')$. We generate $\kb'$ from $\kb$ by removing all weights and transforming each formerly weighted formula $f$ into a disjunction $f|not\ f$, where $not$ stands for default negation and $|$ stands for the disjunction in ASP (so probabilities are ``default probabilities'' in our framework). Note that $f|not\ f$ doesn't guarantee that answer sets are generated for weighted formula $f$. By using ASP choice constructs such as aggregates and disjunctions, the user can basically generate as many answer sets (possible worlds) as desired. 

Formulas do not need to be ground - as defined in Section \ref{syntax}, they can contain existentially as well as universally quantified variables in the FOL sense (although restricted to finite domains).

\noindent As an example, consider the following simple \textit{ground} \PrASP program (examples for \PrASP programs with variables and first-order style quantifiers are presented in the next sections):

{\small \begin{lstlisting}
[0.7] q <- p.
[0.3] p.
[0.2] -p & r.
\end{lstlisting}}

\noindent The set of answer sets (which we take as possible worlds) of the spanning program of this \PrASP program is $\{\{ p, q \}, \{ -p, r \}, \{ \}, \{ p \}\}$.
 
The semantics of a \PrASP program $\kb$ and single \PrASP formulas is defined in terms of a probability distribution over a set of possible worlds (in form of answer sets of $\dmh(\kb)$) in connection with the stable model semantics. This is analogously to the use of \textit{Type 2 probability structures} \cite{halpern90} for first-order probabilistic logics with probabilities, but restricted to finite domains of discourse. 

Let $M = (D, \Theta, \pi, \mu)$ be a probability structure where $D$ is a finite discrete domain of objects, $\Theta$ is a non-empty set of possible worlds, $\pi$ a function which assigns to the symbols in $\varPhi$ (see Section \ref{syntax}) predicates, functions and objects over/from $D$, and $\mu$ a discrete probability function over $\Theta$.\\    
Each possible world is a Herbrand interpretation over $\varPhi$. Since we will use answer sets as possible worlds, defining $\as(a)$ to be the set of all answer sets of answer set program $a$ will become handy. For example, given $\dmh(\kb)$ as (uncertain) knowledge, the set of worlds deemed possible according to existing belief $\dmh(\kb)$ is $\as(\dmh(\kb))$ in our framework.

We define a (non-probabilistic) satisfaction relation of possible worlds and unannotated programs  as follows: let $\kb^-$ be is an unannotated program. Then $(M, \theta) \vDash_\Theta \kb^-$ \textit{iff} $\theta \in \as(\folTolp(\kb^-))$ and $\theta \in \Theta$ (from this it follows that $\Theta$ induces its own closed world assumption - any answer set which is not in $\Theta$ is not satisfiable wrt. $\vDash_\Theta$). The probability $\mu(\{\theta\})$ of a possible world $\theta$ is denoted as $Pr(\theta)$ and sometimes called ``weight'' of $\theta$. For a disjunctive program $\psi$, we analogously define $(M, \theta) \vDash_\Theta \psi$ \textit{iff} $\theta \in \as(\psi)$ and $\theta \in \Theta$.

To do groundwork for the computation of a probability distribution over possible worlds $\Theta$ which are ``generated'' and weighted by some given background knowledge in form of a \PrASP program, we define a (non-probabilistic) satisfaction relation of possible worlds and unannotated formulas: let $\phi$ be a PrASP formula (without weight) and $\theta$ be a possible world. Then $(M, \theta) \vDash_{\kb} \phi$ \textit{iff} $(M, \theta) \vDash_\Theta \dmh(\kb) \cup \folTolp(\phi)$ \textit{and} $\Theta = \as( \dmh(\kb) )$  
(we say formula $\phi$ is \textit{true in possible world} $\theta$). Sometimes we will just write $\theta \models_{\kb} \phi$ if $M$ is given by the context. A notable property of this definition is that it does not restrict us to single ground formulas. Essentially, an unannotated formula $\phi$ can be any answer set program specified in FOL syntax, even if its grounding consists of multiple sentences. Observe that $\Theta$ restricts $\vDash_{\kb}$ to answer sets of $\dmh(\kb)$. For convenience, we will abbreviate $(M, \theta) \vDash_{\kb} \phi$ as $\theta \vDash_{\kb} \phi$.

$Pr(\phi)$ denotes the probability of a formula $\phi$, with $Pr(\phi) = \mu(\{ \theta \in \Theta: (M, \theta) \vDash_{\kb} \phi\})$. Note that this holds both for annotated and unannotated formulas: even if it has a weight attached, the probability of a \PrASP formula is defined by means of $\mu$ and only indirectly by its manually assigned weight (weights are used below as constraints for the computation of a probabilistically consistent $\mu$). Further observe that there is no particular treatment for conditional probabilities in our framework; $Pr(a|b)$ is simply calculated as $Pr(a \wedge b) / Pr(b)$.\\
While our framework so far is general enough to account for probabilistic inference using unrestricted  programs and query formulas (provided we are given a probability distribution over the possible answer sets), this generality also means a relatively high complexity in terms of computability for inference-heavy tasks which rely on the repeated application of operator $\vDash_{\kb}$, even if we would avoid the transformation $\folTolp$ and restrict ourselves to the use of ASP syntax. 

The obvious question now, addressed before for other probabilistic logics, is how to compute $\mu$, i.e., how to obtain a probability distribution over possible worlds (which tells us for each possible world the probability with which this possible world is the actual world) from a given annotated  program $\kb$ in a sound and computationally inexpensive way.\\
Generally, we can express the search for probability distributions in form of a number of constraints which constitute a system of linear inequalities (which reduce to linear equalities for point probabilities as weights). This system typically has multiple or even infinitely many solutions (even though we do not allow for probability intervals) and computation can be costly, depending on the number of possible worlds according to $ \dmh(\kb)$.\\
We define the parameterized probability distribution $\mu(\kb, \Theta)$ over a set $\Theta$ of answer sets as the solution (for all $Pr(\theta_i)$) of the following system of linear equations and an inequality (if precisely one solution exists) or as the solution with maximum entropy \cite{journals/igpl/ThimmK12}, in case multiple solutions exist \footnote{Since in this case the number of solutions of the system of linear equations is infinite, de facto we need to choose the maximum entropy solution of some finite subset. In the current prototype implementation, we generate a user-defined number of random solutions derived from a solution computed using a constrained variant of Singular Value Decomposition and the null space of the coefficient matrix of the system of linear equations (1)-(3).}. We require that the given weights in a PrASP program are chosen such that the following constraint system has at least one solution. 
{\small \begin{gather}
\label{linearSystem}
 \sum_{\theta_i \in \Theta: \theta_i \vDash_{\kb} f_1}{Pr(\theta_i)}   = w(f_1) 
\end{gather}
\vspace{-0.2cm}
\centerline{$ \cdots$}
\vspace{-0.2cm}
\begin{gather}
\sum_{\theta_i \in \Theta: \theta_i \vDash_{\kb} f_n}{Pr(\theta_i)}   = w(f_n)\\
\sum_{\theta_i \in \Theta}{\theta_i = 1}\\
\forall \theta_i \in \Theta: 0 \leq Pr(\theta_i) \leq 1 
\label{linearSystem4}
\end{gather}}

\noindent At this, $\kb = \{ f_1, ..., f_n \}$ is a PrASP program.

The \textit{canonical probability distribution} $\mu(\kb)$ of $\kb$ is defined as $\mu(\kb, \as(\dmh(\kb)))$. In the rest of the paper, we refer to $\mu(\kb)$ when we refer to the probability distribution over the answer sets of the spanning program of a given PrASP program $\kb$.

\section{Inference}
\label{Sampling}

Given possible world weights ($\mu(\kb)$), probabilistic inference becomes a model counting task where each model has a weight: we can compute the probability of any query formula $\phi$ by summing up the probabilities (weights) of those possible worlds (models) where $\phi$ is true. To make this viable even for larger sets of possible worlds, we optionally restrict the calculation of $\mu(\kb)$ to a number of answer sets sampled near-uniformly at random from the total set of answer sets of the spanning program, as described in Section \ref{xor}. 

\subsection{Adding a sampling step and computing probabilities}
\label{xor}

All tasks described so far (solving the system of (in)equalities, counting of weighted answer sets) become intractable for very large sets of possible worlds. To tackle this issue, we want to restrict the application of these tasks to a sampled subset of all possible worlds. Concretely, we want to find a way to sample (near-)uniformly from the total set of answer sets \textit{without} computing a very large number of answer sets. While this way the set of answer sets cannot be computed using only a single call of the ASP solver but requires a number of separate calls (each with different sampling constraints), the required solver calls can be performed \textit{in parallel}. However, a shortcoming of the sampling approach is that there is currently no way to pre-compute the size of the minimally required set of samples.

Guaranteeing near-uniformity in answer set sampling looks like a highly non-trivial task, since any set of answers obtained from ASP solvers as a subset of the total set of answer sets is typically not uniformly distributed but strongly biased in hardly foreseeable ways (due to various interplaying heuristics applied by modern solvers), so we could not simply request any single answer set from the solver. 

However, we can make use of so-called \textit{XOR constraints} (a form of streamlining constraints in the area of SAT solving) for near-uniform sampling \cite{DBLP:conf/nips/GomesSS06} to obtain samples from the space of all answer sets, within arbitrarily narrow probabilistic bounds, using any off-the-shelf ASP solver. Compared to approaches which use Markov Chain Monte Carlo (MCMC) methods to sample from some given distribution, this method has the advantage that the sampling process is typically faster and that it requires only an off-the-shelf ASP solver (which is in the ideal case employed only once per sample, in order to obtain a single answer set). However, a shortcoming is that we are not doing Importance Sampling this way - the probability of a possible world is not taken into account but computed later from the samples.\\
Counting answer sets could also be achieved using XOR constraints, however, this is not covered in this paper, since it does not comprise weighted counting, and we could normally not use an unweighted counting approach directly.

XOR constraints were originally defined over a set of propositional variables, which we identify with a set of ground atoms $V = \{a_1, ..., a_n\}$. Each XOR constraint is represented by a subset $D$ of $V \cup \{ true \}$. $D$  is satisfied by some model if an \textit{odd} number of elements of $D$ are satisfied by this model (i.e., the constraint acts like a parity of $D$). In ASP syntax, an XOR constraint can be represented for example as \texttt{:- \#even\{ $a_1$, ..., $a_n$ \}} \cite{Gebser:2011:PPA:1971622.1971623}. \\
In our approach, XOR constraints are independently at random drawn from a probability distribution $\mathbb{X}(|V|, 0.5)$ over the set $V$ of all possible XOR constraints over all ground atoms of the ground answer set program resulting from $\dmh(\kb)$. $\mathbb{X}(|V|, 0.5)$ is defined such that each XOR constraint is drawn from this distribution independently at random with probability 0.5 and includes $true$ with probability 0.5. In effect, any given XOR constraint is drawn with probability $2^{-(|V|+1|)}$ (see \cite{DBLP:conf/nips/GomesSS06} for details). Since adding an XOR constraint to an answer set program eliminates any given answer set with probability 0.5, it cuts the set of answer sets in half in expectation. Iteratively adding a small number of XOR constraints to an answer set program therefore reduces the number of answer sets to a small number also. If this process results in a single answer set, the remaining answer set is drawn near-uniformly from the original set of answer sets, as shown in \cite{DBLP:conf/nips/GomesSS06}.\\
Since for answer set programs the costs of repeating the addition of constraints until precisely a single answer set remains appears to be higher than the costs of computing somewhat too many models, we just estimate the number of required constraints and choose randomly from the resulting set of answer sets. The following way of answer set sampling using XOR constraints has been used before in Xorro (a tool which is part of the \textit{Potassco} set of ASP tools \cite{Gebser:2011:PPA:1971622.1971623}) in a very similar way.\\ 

\noindent Function \textbf{\textit{sample}}: $\psi \mapsto \gamma$
\begin{algorithmic}
\STATE Given any disjunctive program $\psi$, the following procedure computes a random sample $\gamma$ from the set of all answer sets of $\psi$:
\STATE $\psi_g \leftarrow$ ground($\psi$)
\STATE $ga \leftarrow atoms(\psi_g)$
\STATE $xors \leftarrow $ XOR constraints $\{ xor_1, ..., xor_n \}$ over $ga$, drawn from $\mathbb{X}(|V|, 0.5)$
\STATE $\psi' \leftarrow \psi \cup xors$
\STATE $\gamma \leftarrow $ an answer set selected randomly from $\as(\psi')$\\
\STATE
\end{algorithmic}
\noindent At this, the number of constraints $n$ is set to a value large enough to produce one or a very low number of answer sets ($log_2(|ga|)$ in our experiments). \\

We can now compute $\mu(\kb, \Theta')$ (i.e., $Pr(\theta)\ \mathrm{for\ each\ } \theta \in \Theta'$) for a set of samples $\Theta'$ obtained by multiple (ideally parallel) calls of \textit{sample} from the spanning program $\dmh(\kb)$ of PrASP program $\kb$, and subsequently sum up the weights of those samples (possible worlds) where the respective query formula (whose marginal probability we want to compute) is true. Precisely, we approximate $Pr(\phi)$ for a (ground or non-ground) query formula $\phi$ using: 
{\small \begin{align}
Pr(\phi) \approx 
\sum_{\{\theta' \in \Theta': \theta' \models_{\kb} \phi\}} {Pr(\theta')} 
\end{align}}
\noindent for a sufficiently large set $\Theta'$ of samples. 

\noindent Conditional probabilities $Pr(a|b)$ can simply be computed as $Pr(a \wedge b) / Pr(b)$.

If sampling is not useful (i.e., if the total number of answer sets $\Theta$ is moderate), inference is done in the same way, we just set $\Theta' = \Theta$. Sampling using XOR constraints costs time too (mainly because of repeated calls of the ASP solver), and making this approach more efficient is an important aspect of future work (see Section \ref{conclusions}). \\

\noindent As an example for inference using our current implementation, consider the following PrASP formalization of a simple coin game:
{\small \begin{verbatim}
coin(1..3).
[0.6] coin_out(1,heads).
[[0.5]] coin_out(N,heads) :- coin(N), N != 1.
1{coin_out(N,heads), coin_out(N,tails)}1 
         :- coin(N).
n_win :- coin_out(N,tails), coin(N).
win :- not n_win. 
\end{verbatim}}
At this, the line starting with \verb|[[0.5]]...| is syntactic sugar for a set of weighted rules where variable N is instantiated with all its possible values (i.e.,\\
 {\small \verb|[0.5] coin_out(2,heads) :- coin(2), 2 != 1|} and \\
  {\small \verb|[0.5] coin_out(3,heads) :- coin(3), 3 != 1|}). It would also be possible to use \verb|[0.5]| as annotation of this rule, in which case the weight 0.5 would specify the probability of the whole non-ground formula instead.\\ 
Our prototypical implementation accepts query formulas in format \verb|[?] a|$\ $ (computes the marginal probability of a) and \verb#[?|b] a#$\ $ (computes the conditional probability $Pr(a|b)$). E.g.,
{\small \begin{verbatim}
[?] coin_out(1,tails).
[?] coin_out(1,heads) | coin_out(1,tails).
[?] coin_out(1,heads) & coin_out(2,heads) 
        & coin_out(3,heads).
[?] win.
[?|coin_out(1,heads) & coin_out(2,heads) 
         coin_out(3,heads)] win.
\end{verbatim} } 
\noindent ...yields the following result
{\small \begin{verbatim}
[0.3999999999999999] coin_out(1,tails).
[1] coin_out(1,heads) | coin_out(1,tails).
[0.15] coin_out(1,heads) & coin_out(2,heads) 
       & coin_out(3,heads).
[0.15] win.
[1|coin_out(1,heads) & coin_out(2,heads) 
   & coin_out(3,heads)] win.
\end{verbatim} }
In this example, use of sampling does not make any difference due to its small size. An example where a difference can be observed is presented in Section \ref{weightLearning}. This example also demonstrates that FOL and logic programming / ASP syntax can be freely mixed in background knowledge and queries.\\
Another simple example shows the use of FOL-style variables and quantifiers mixed with ASP-style variables:

\begin{verbatim}
p(1). p(2). p(3).
#domain p(X).
[0.5] v(1). 
[0.5] v(2).
[0.5] v(3).
[0.1] v(X).
\end{verbatim}

\noindent With this, the following query:

\begin{verbatim}
[?] v(X).
#domain p(Z).
[?] ![Z]: v(Z).
[?] ?[Z]: v(Z).
\end{verbatim}

\noindent ...results in: 

\begin{verbatim}
[0.1] ![Z]: v(Z).
[0.8499999999999989] ?[Z]: v(Z).
\end{verbatim}

The result of query \verb|[?] ![Z]: v(Z)| with universal quantifier \verb|![Z]| is $Pr(\forall z.v(z)) = 0.1$, which is also the result of the  equivalent queries \verb|[?] v(1) & v(2) & v(3)| and \verb|[?] v(X)|. In our example, this marginal probability was directly given as weight in the background knowledge. In contrast to \verb|X|, variable \verb|Z| is a variable in the sense of first-order logic (over a finite domain). \\
The result of \verb|?[Z]: v(Z)| is $Pr(\exists z.v(z))$ (i.e., \verb|?[Z]:| represents the existential quantifier) and could likewise be calculated manually using the inclusion-exclusion principle as $Pr(v(1)\vee v(2) \vee v(3)) = Pr(v(1)) + Pr(v(2)) + Pr(v(3)) - Pr(v(1) \wedge v(2)) - Pr(v(1) \wedge v(3)) - Pr(v(2) \wedge v(3)) + Pr(v(1) \wedge v(2) \wedge v(3)) = 0.85$.\\
Of course, existential or universal quantifiers can also be used as sub-formulas and in PrASP programs.

\subsection{An alternative approach: conversion into an equivalent non-probabilistic answer set program} 
\label{choiceConstructs}

An alternative approach to probabilistic inference without computing $\mu$ and without counting of weighted possible worlds, would be to find an unannotated first-order program $\kb'$ which reflects the desired probabilistic nondeterminism (choice) of a given PrASP program $\kb$. Instead of defining probabilities of possible worlds, $\kb'$ has answers sets whose frequency (number of occurrences within the total set of answer sets) reflects the given probabilities in the original (annotated) program. To make this idea more intuitive, imagine that each possible world corresponds to a room. Instead of encountering a certain room with a certain frequency, we create further rooms which have all, from the viewpoint of the observer, the same look, size and furniture. The number of these rooms reflects the probability of this type of room. E.g., to ensure probability $\frac{1}{3}$ of some literal $p$, $\kb'$ is created in a way such that $p$ holds in one third of all answer sets of $\kb'$. This task can be considered as an elaborate variant of the generation of the (much simpler) spanning program $\dmh(\kb)$.

Finding $\kb'$ could be formulated as an (intractable) rule search problem (plus subsequently the conversion into ASP syntax and a simple unweighted model counting task): find a non-probabilistic program $\kb'$ such that for each annotated formula $[p] f$ in the original program the following holds (under the provision that the given weights are probabilistically sound):
{\small \begin{gather}
\frac{|\{ m: m \in \as(\kb'), m \models f \}|}{| \as(\kb')|} = p.
\end{gather} }
Unfortunately, the direct search approach to this would be obviously intractable.

However, in the special case of mutually independent formulas we can omit the rule learning task by conditioning each formula in $\kb$ by a nondeterministic choice amongst the truth conditions of a number of ``helper atoms'' $h_i$ (which will later be ignored when we count the resulting answer sets), in order to ``emulate'' the respective probability specified by the weight. If (and only if) the formulas are mutually independent, the obtained $\kb'$ is isomorphic to the original probabilistic program. In detail, conditioning means to replace each formula $[w]\ f$ by formulas $1\{ h_1, ..., h_n\}1$, $f \leftarrow h_1|...|h_m$ and $not\ f \leftarrow not\ (h_1|...|h_m)$, where the $h_i$ are new names (the aforementioned ``helper atoms''), $\frac{m}{n} = w$ and $m < n$ (remember that we allow for weight constraints as well as FOL syntax).\\ 

In case the transformation accurately reflects the original uncertain program, we could now calculate marginal probabilities simply by determining the percentage of those answer sets in which the respective query formula is true (ignoring any helper atoms introduced in the conversion step), with no need for computing $\mu(\kb)$. 

\noindent As an example, consider the following program:
{\footnotesize \begin{verbatim}
coin(1..10).
[0.6] coin_out(1,heads).
[[0.5]] coin_out(N,heads) :- coin(N), N != 1.

1{coin_out(N,heads), coin_out(N,tails)}1 
  :- coin(N).
n_win :- coin_out(N,tails), coin(N).
win :- not n_win.
\end{verbatim}}
Since coin tosses are mutually independent, we can transform it into the following equivalent un-annotated form (the \texttt{hpatom}$_n$ are the ``helper atoms''. Rules are written as disjunctions):

{\footnotesize \begin{verbatim}
coin(1..10).
1{hpatom1,hpatom2,hpatom3,hpatom4,hpatom5}1.
(coin_out(1,heads))
 | -(hpatom1|hpatom2|hpatom3).
not (coin_out(1,heads)) 
 | (hpatom1|hpatom2|hpatom3).
1{hpatom6,hpatom7}1.
(coin_out(10,heads)) | -(hpatom6).
not (coin_out(10,heads)) | (hpatom6).
1{hpatom8,hpatom9}1.
(coin_out(9,heads)) | -(hpatom8).
not (coin_out(9,heads)) | (hpatom8).
1{hpatom10,hpatom11}1.
(coin_out(8,heads)) | -(hpatom10).
not (coin_out(8,heads)) | (hpatom10).
1{hpatom12,hpatom13}1.
(coin_out(7,heads)) | -(hpatom12).
not (coin_out(7,heads)) | (hpatom12).
1{hpatom14,hpatom15}1.
(coin_out(6,heads)) | -(hpatom14).
not (coin_out(6,heads)) | (hpatom14).
1{hpatom16,hpatom17}1.
(coin_out(5,heads)) | -(hpatom16).
not (coin_out(5,heads)) | (hpatom16).
1{hpatom18,hpatom19}1.
(coin_out(4,heads)) | -(hpatom18).
not (coin_out(4,heads)) | (hpatom18).
1{hpatom20,hpatom21}1.
(coin_out(3,heads)) | -(hpatom20).
not (coin_out(3,heads)) | (hpatom20).
1{hpatom22,hpatom23}1.
(coin_out(2,heads)) | -(hpatom22).
not (coin_out(2,heads)) | (hpatom22).
1{coin_out(N,heads), coin_out(N,tails)}1 
  :- coin(N).
n_win :- coin_out(N,tails), coin(N).
win :- not n_win.
\end{verbatim}}

\noindent Exemplary query results:

{\footnotesize \begin{verbatim}
[0.001171875] win.
[0.998828125] not win.
[0.6] coin_out(1,heads).
[0.5] coin_out(2,heads).
\end{verbatim}}

What is remarkable here is that no equation solving task (computation of $\mu(\kb)$) is required to compute these results. However, this does not normally lead to improved inference speed, due to the larger amount of time required for the computation of models.

\section{Weight Learning}
\label{weightLearning}

Generally, the task of parameter learning in probabilistic inductive logic programming is to find probabilistic parameters (weights) of logical formulas which maximize the likelihood given some data (learning examples) \cite{DBLP:conf/ilp/RaedtK08}. In our case, the hypothesis $H$ (a set of formulas without weights) is provided by an expert, optionally together with some PrASP program as background knowledge $B$. The goal is then to discover weights $w$ of the formulas $H$ such that $Pr(E | H_w \cup B)$ is maximized given example formulas $E = {e_1, e_2,...}$. Formally, we want to compute
{\small \begin{gather}
\label{inductiveProbTask}
argmax_w(Pr(E | H_w \cup B)) = argmax_w(\prod_{e_i \in E}Pr(e_i | H_w \cup B))
\end{gather}}
(Making the usual i.i.d. assumption regarding the individual examples in $E$. $H_w$ denotes the hypothesis weighted with weight vector $w$.) 

This results in an optimization task which is related but not identical to weight learning for, e.g., MLNs and \cite{Corapi:2011:PRL:2044543.2044565}. In MLNs, typically a database (possible world) is given whose likelihood should be maximized, e.g. using a generative approach \cite{Lowd07efficientweight} by gradient descent. Another related approach distinguishes a priori between evidence atoms $X$ and query atoms $Y$ and seeks to maximize the likelihood $Pr(Y|X)$, again using gradient descent \cite{Huynh08discriminativestructure}. At this, cost-heavy inference is avoided as far as possible, e.g., by optimization of the pseudo-(log-)likelihood instead ot the (log-)likelihood or by approximations of costly counts of true formula groundings in a certain possible world (the basic computation in MLN inference). In contrast, the current implementation of PrASP learns weights from any formulas and not just literals (or, more precisely as for MLNs: atoms, where negation is implicit using a closed-world assumption). Furthermore, the maximization targets are different ($Pr(\mathit{possible\ world})$ or $Pr(Y|X)$) vs. $Pr(E | H_w \cup B)$). 

Regarding the need to reduce inference when learning, PrASP parameter estimation should in principle make no exception, since inference can still be costly even when probabilities are inferred only approximately by use of sampling. However, in our preliminary experiments we found that at least in relatively simple scenarios, there is no need to resort to inference-free approximations such as pseudo-(log-)likelihood. The pseudo-(log-)likelihood approach presented in early works on MLNs \cite{mln} would also require a probabilistic ground formula independence analysis in our case, since in PrASP there is no obvious equivalent to Markov blankets.\\
Note that we assume that the example data is non-probabilistic and fully observable.

Let $H = \{f_1, ..., f_n\}$ be a given set of formulas and a vector $w = (w^1,...,w^n)$ of (unknown) weights of these formulas. Using the Barzilai and Borwein method \cite{barzilai} (a variant of the gradient descent approach with possibly superlinear convergence), we seek to find $w$ such that $Pr(E | H_w \cup B)$ is maximized ($H_w$ denotes the formulas in $H$ with the weights $w$ such that each $f_i$ is weighted with $w^i$). 
Any existing weights of formulas in the background knowledge ar not touched, which can significantly reduce learning complexity if $H$ is comparatively small. Probabilistic or unobservable examples are not considered.

\noindent The learning algorithm \cite{barzilai} is as follows:\\

\noindent Repeat for $k=0,1,...$ until convergence:\\
\indent Set $s_k = \frac{1}{\alpha_k} \triangledown(Pr(E|H_{w_k} \cup B))$\\
\indent Set $w_{k+1} = w_{k} + s_k$\\
\indent Set $y_{k} = \triangledown(Pr(E|H_{w_{k+1}} \cup B)) - \triangledown(Pr(E|H_{w_k} \cup B))$\\

\indent Set $\alpha_{k+1} = \frac{s^T_k y_k}{s^T_k s_k}$\\

At this, the initial gradient ascent step size $\alpha_0$ and the initial weight vector $w_0$ can be chosen freely. 
$Pr(E|H_w \cup B)$ denotes $\prod_{e_i \in E}Pr(e_i | H_w \cup B)$ inferred using vector $w$ as weights for the hypothesis formulas, and
{\small \begin{gather}
\bigtriangledown(Pr(E|H_w \cup B)) =\\
	(\frac{\partial}{\partial w^1} Pr(E|H_w \cup B), ..., \frac{\partial}{\partial w^n} Pr(E|H_w \cup B))
\end{gather}}
 
Since we usually cannot practically express $Pr(E|H_w \cup B)$ in dependency of $w$ in closed form, at a first glance, the above formalization appears to be not very helpful. However, we can still resort to numerical differentiation and approximate
{\small \begin{gather}
\bigtriangledown(Pr(E|H_w \cup B)) =\\
(\lim_{h \rightarrow 0} \frac{Pr(E|H_{(w^1+h, ..., w^n)} \cup B)-Pr(E|H_{(w^1, ..., w^n)} \cup B)}{h},
\end{gather}
\centerline{{\large ...,}}
\begin{gather}
\lim_{h \rightarrow 0} \frac{Pr(E|H_{(w^1, ..., w^n+h)} \cup B)-Pr(E|H_{(w^1, ..., w^n)} \cup B)}{h})
\end{gather}}
by computing the above vector (dropping the limit operator) for a sufficiently small $h$ (in our prototypical implementation, $h = \sqrt{\epsilon}w_i$ is used, where $\epsilon$ is an upper bound to the rounding error using the machine's double-precision floating point arithmetic).\\
This approach has the benefit of allowing in principle for any maximization target (not just $E$). In particular, any unweighted formulas (unnegated and negated facts as well as rules) can be used as (positive) examples.\\

As a small example both for inference and weight learning using our preliminary implementation, consider the following fragment of a an nonmonotonic indoor localization scenario, which consists of estimating the position of a person, and determining how this person moves a certain number of steps around the environment until a safe position is reached: 

{\small \begin{verbatim}
[0.6] moved(1).
[0.2] moved(2).
point(1..100).
1{atpoint(X):point(X)}1.
distance(1) :- moved(1).
distance(2) :- moved(2).
atpoint(29) | atpoint(30) | atpoint(31) 
   | atpoint(32) | atpoint(33) 
   | atpoint(34) | atpoint(35) | atpoint(36) 
   | atpoint(37) -> selected.
safe :- selected, not exception.
exception :- distance(1).
\end{verbatim}}

The spanning program of this example has 400 answer sets. Inference of \\
$Pr(\mathit{safe}|\mathit{distance}(2))$ and $Pr(\mathit{safe}|\mathit{distance}(1))$ without sampling requires ca. 2250 ms using our current unoptimized prototype implementation. If we increase the number of points to 1000, inference is tractable only by use of sampling (see Section \ref{Sampling}). \\
To demonstrate how the probability of a certain hypothesis can be learned in this simple scenario, we remove \verb|[0.6] moved(1)| from the program above (with 100 points) and turn this formula (without the weight annotation) into a hypothesis. Given example data \verb|safe|, parameter estimation results in $Pr(moved(1)) \approx 0$, learned in ca. 3170 ms using our current prototype implementation. 

\section{Conclusions}
\label{conclusions}

With this introductory paper, we have presented a novel framework for uncertainty reasoning and parameter estimation based on Answer Set Programming, with support for probabilistically weighted formulas in background knowledge, hypotheses and queries. While our current framework certainly leaves room for future improvements, we believe that we have already pointed out a new venue towards more practicable probabilistic inductive answer set programming with a high degree of expressiveness. Ongoing work is focusing on performance improvements, theoretical analysis (in particular regarding minimum number of samples wrt. inference accuracy), empirical evaluation and on the investigation of viable approaches to PrASP structure learning.\\

\noindent\textbf{Acknowledgments}

\noindent This work is supported by the EU FP7 CityPulse Project under grant No. 603095. http://www.ict-citypulse.eu

\bibliographystyle{aaai}
\bibliography{nmr14}

\end{document}